\documentclass{article}

\PassOptionsToPackage{numbers, compress}{natbib}
\usepackage[preprint]{neurips_2019}
\usepackage[utf8]{inputenc} 
\usepackage[T1]{fontenc}    
\usepackage{hyperref}       
\usepackage{url}            
\usepackage{booktabs}       
\usepackage{amsfonts}       
\usepackage{nicefrac}       
\usepackage{microtype}      
\usepackage{lipsum}
\usepackage{graphicx} 
\usepackage{amssymb}
\usepackage{amsthm}
\usepackage{amsmath} 
\usepackage{algorithmic}
\usepackage{algorithm}
\usepackage{color}
\usepackage{paralist}
\usepackage{arydshln}
\usepackage{multirow}
\usepackage{bbm}
\usepackage{xspace}

\title{Temporal Collaborative Ranking \\ Via Personalized Transformer}

\author{%
  Liwei Wu \\
  Department of Statistics\\
  University of California, Davis\\
  Davis, CA 95616 \\
  \texttt{liwu@ucdavis.edu} \\
  \And
  Shuqing Li \\
  Department of Computer Science\\
  University of California, Davis\\
  Davis, CA 95616 \\
  \texttt{qshli@ucdavis.edu}
  \And
  Cho-Jui Hsieh \\ 
  Department of Computer Science\\
  University of California, Los Angles\\
  Los Angles, CA 90095 \\
  \texttt{chohsieh@cs.ucla.edu}
  \And
  James Sharpnack \\
  Department of Statistics\\
  University of California, Davis\\
  Davis, CA 95616 \\
  \texttt{jsharpna@ucdavis.edu} \\
}

\begin{document}
\maketitle

\begin{abstract}
The collaborative ranking problem has been an important open research question as most recommendation problems can be naturally formulated as ranking problems. While much of collaborative ranking methodology assumes static ranking data, the importance of temporal information to improving ranking performance is increasingly apparent. Recent advances in deep learning, especially the discovery of various attention mechanisms and newer architectures in addition to widely used RNN and CNN in natural language processing, have allowed us to make better use of the temporal ordering of items that each user has engaged with. In particular, the SASRec model, inspired by the popular Transformer model in natural languages processing, has achieved state-of-art results in the temporal collaborative ranking problem and enjoyed more than 10x speed-up when compared to earlier CNN/RNN-based methods. However, SASRec is inherently an un-personalized model and does not include personalized user embeddings. To overcome this limitation, we propose a Personalized Transformer (SSE-PT) model, outperforming SASRec by almost 5\% in terms of NDCG@10 on 5 real-world datasets. Furthermore, after examining some random users' engagement history and corresponding attention heat maps used during the inference stage, we find our model is not only more interpretable but also able to focus on recent engagement patterns for each user. Moreover, our SSE-PT model with a slight modification, which we call SSE-PT++, can handle extremely long sequences and outperform SASRec in ranking results with comparable training speed, striking a balance between performance and speed requirements. Code and data are open sourced at \url{https://github.com/wuliwei9278/SSE-PT}.
\end{abstract}

\section{Introduction}
Recommendation systems are increasingly prevalent due to content delivery platforms, e-commerce websites, and mobile apps \cite{shani2008mining}. Most recommendation problems can be naturally thought of as predicting the user's partial ranking of a large candidate pool of items.  After obtaining the optimal ranking ordering, the recommender system can simply recommend top-$K$ items in the list for each individual user. Usually rankings are made personalized to cater to users' special tastes. In literature, this is formulated as the collaborative ranking problem \cite{weimer2008cofi}. The temporal ordering, determined by when users engaged with items, has proven to be an important resource to further improve ranking performance. We call the collaborative ranking setting with temporal ordering information the Temporal Collaborative Ranking problem in this paper.

Recent advances in deep learning, especially the discovery of various attention mechanisms \cite{bahdanau2014neural, sutskever2014sequence} and newer architectures \cite{vaswani2017attention, devlin2018bert} in addition to classical RNN and CNN architecture in natural language processing, have allowed us to make better use of the temporal ordering of items that each user has engaged with. In particular, the SASRec model \cite{kang2018self}, inspired by the popular Transformer model in natural languages processing, has achieved state-of-art results in the temporal collaborative ranking problem and enjoyed more than 10x speed-up  compared to earlier RNN\cite{hidasi2015session} / CNN\cite{tang2018personalized}-based methods. But at a closer look, the SASRec is inherently an un-personalized model without introducing user embeddings and this often leads to an inferior recommendation model in terms of both ranking performances and model interpretability. Although personalization is not needed for the original Transformer model \cite{vaswani2017attention} in natural languages understandings or translations, personalization plays a crucial role throughout recommender system literature \cite{zhang2019deep} ever since the matrix factorization approach to the Netflix prize \cite{koren2009bellkor}.

In this work, we propose a novel method, Personalized Transformer (SSE-PT), that introduces personalization into self-attentive neural network architectures. 
\cite{kang2018self} found that adding additional personalized embeddings did not improve the performance of their Transformer model, and postulate that this is due to the fact that they already use the user history and the embeddings only contribute to overfitting.
Although introducing user embeddings into the model is indeed difficult with existing regularization techniques for embeddings, we show that personalization can greatly improve ranking performances with recent regularization technique called Stochastic Shared Embeddings (SSE) \cite{wu2019stochastic}. 
The personalized Transformer (SSE-PT) model with SSE regularization works well for all 5 real-world datasets we consider, outperforming previous state-of-the-art algorithm SASRec by almost 5\% in terms of NDCG@10. Furthermore, after examining some random users' engagement history and corresponding attention heat maps used during the inference stage, we find our model is not only more interpretable but also able to focus on recent engagement patterns for each user. Moreover, our SSE-PT model with a slight modification, which we call SSE-PT++, can handle extremely long sequences and outperform SASRec in ranking results with comparable training speed, striking a balance between performance and speed requirements.


\section{Related Work}
\subsection{Collaborative Filtering and Ranking}
Recommender systems can be divided into those designed for explicit feedback, such as ratings \cite{koren2009matrix}, and those for implicit feedback, based on user engagement \cite{hu2008collaborative}. 
Recently, implicit feedback datasets, such as user clicks of web pages, check-in’s of restaurants, likes of posts, listening behavior of music, watching history and purchase history, are increasingly prevalent. 
Unlike explicit feedback, implicit feedback datasets can be obtained without users noticing or active participation. 
Item-to-item \cite{sarwar2001item}, user-to-user \cite{wang2006unifying}, user-to-item \cite{koren2009matrix} are 3 different angles of utilizing user engagement data. In item-to-item approaches, the goal is to recommend similar items to what users have engaged. In user-to-user approaches, the goal is to recommend to a user some items that similar users have engaged previously. User-to-item approaches, on the other hand, focus on examining user and item relationships as a whole, which is also referred to as a collaborative filtering approach. These relationships can also be viewed as graphs \cite{wu2019graph}.

Two main approaches to recommendations are: attempt to predict the explicit or implicit feedback with matrix (or tensor) completion, or attempt to predict the relative rankings derived from the feedback. 
Collaborative filtering algorithms including matrix factorization,  \cite{hill1995recommending,schafer2007collaborative,koren2008factorization,mnih2008probabilistic,hu2008collaborative}, which predict the feedback in a pointwise fashion as if it were a supervised learning problem, fall into the first category.
Predicting the feedback with supervised learning objectives suffers from the different rating standards of users, and it can be helpful to consider the data to simply be the ranking of the items based on feedback.
There are two main approaches to the collaborative ranking problem, namely pairwise and listwise methods. 
Pairwise methods \cite{rendle2009bpr, wu2017large} consider each pairwise comparison for a user as a label, which implicitly models the pairwise comparisons as independent observations. 
Listwise methods \cite{wu2018sql}, on the other hand, consider a user's entire engagement history as independent observations. Normally, in terms of ranking performances, listwise approaches outperform pairwise approaches, and pairwise approaches outperform pointwise collaborative filtering \cite{wu2018sql}. 


\subsection{Session-based and Sequential Recommendation}
Both session-based and sequential (i.e. next-basket) recommendation algorithms take advantage of additional temporal information to make better personalized recommendations. The main difference between session-based recommendations \cite{hidasi2015session} and sequential recommendations \cite{kang2018self} is that the former assumes that the user ids are not recorded and therefore the length of engagement sequences are relatively short. Therefore, session-based recommendations normally do not consider user factors. On the other hand, sequential recommendation treats each sequence as a user's engagement history \cite{kang2018self}.
Both settings, do not explicitly require time-stamps: only the relative temporal orderings are assumed known (in contrast to, for example, timeSVD++ \cite{koren2009collaborative}).
Initially, sequence data in temporal order are usually modelled with Markov models, in which future observation is conditioned on last few observed items \cite{rendle2010factorizing}. 
In \cite{rendle2010factorizing}, a personalized Markov model with user latent factors is proposed for more personalized results. 

In recent years, deep learning techniques, borrowing from natural language processing (NLP) literature, are more widely used in tackling sequential data. Like sentences in NLP, sequence data in recommendations can be similarly modelled by recurrent neural networks (RNN) \cite{hidasi2015session, hidasi2018recurrent} and convolutional neural network (CNN) \cite{tang2018personalized} models. Later on, attention models are getting more and more attention in both NLP, \cite{vaswani2017attention, devlin2018bert}, and recommender systems, \cite{liu2018stamp, kang2018self}. SASRec \cite{kang2018self} is a recent method with state-of-the-art performance among the many deep learning models. 
Motivated by the Transformer model in neural machine translation \cite{vaswani2017attention}, SASRec utilizes a similar architecture to the encoder part of the Transformer model. 


\subsection{Regularization Techniques}
In deep learning, models with many more parameters than data points can easily overfit training data. This may prevent us from adding user embeddings as additional parameters into complicated models like the Transformer model \cite{kang2018self}, which can easily have 20 layers with 6 self-attention blocks and millions of parameters for a medium-sized dataset like Movielens10M \cite{harper2016movielens}. 
$\ell_2$ regularization~\cite{hoerl1970ridge} is the most widely used approach and has been used in many matrix factorization models in recommender systems; $\ell_1$ regularization~\cite{tibshirani1996regression} is used when a sparse model is preferred. For deep neural networks, it has been shown that $\ell_p$  regularizations are often too weak, while dropout~\cite{hinton2012improving,srivastava2014dropout} is more effective in practice.  There are many other regularization techniques, including parameter sharing \cite{goodfellow2016deep}, max-norm regularization \cite{srebro2005maximum}, gradient clipping \cite{pascanu2013difficulty}, etc.
Very recently, a new regularization technique called Stochastic Shared Embeddings (SSE) \cite{wu2019stochastic} is proposed as a new means of regularizing embedding layers. \cite{wu2019stochastic} develops two versions of SSE, SSE-Graph and SSE-SE. We find that SSE-SE is essential to the success of our Personalized Transformer (SSE-PT) model.


\section{Methodology}
\subsection{Temporal Collaborative Ranking}
Let us formally define the temporal collaborative ranking problem as: given $n$ users, each user engaging with a subset of $m$ items in a temporal order, the goal of the task is to 
find an optimal personalized ranking ordering of top $K$ items out of total $m$ items for any given user at any given time point. 
We assume our data is in the format of sequences of items for $n$ users have interacted with so far, namely 
\begin{equation}\label{eq:input}
  s_i=(j_{i1}, j_{i2}, \dots, j_{iT}) \text{ for } 1 \leq i \leq n.  
\end{equation}
 Sequences $s_i$ of length $T$ contain indices of the last $T$ items the user $i$ has interacted with in the temporal order (from old to new). 
For different users, the sequence lengths can be very different (where we pad the shorter sequences to obtain length $T$). We cannot simply randomly split data points into train/validation/test sets because they come in temporal orders. Instead, we need to make sure our training data is before validation data which is before test data temporally.
We use last items in sequences as test sets, second-to-last items as validation sets and the rest as training sets.  We use ranking metrics such as NDCG@$K$ and Recall$@K$ for evaluations, which are defined in \eqref{eq:ndcg} and \eqref{eq:recall}. 

\subsection{Personalized Transformer Architecture}
Our model is motivated by the Transformer model in \cite{vaswani2017attention} and \cite{kang2018self}.
In the following sections, we are going to examine each component of our Personalized Transformer (SSE-PT) model: the embedding layer, self-attention layer, pointwise feed-forward layer, prediction layer, layer normalization, dropout, weight decay, stochastic shared embeddings, and so on. 

\subsubsection{Embedding Layer}
We define a learnable user embedding look-up table $U \in R^{n \times d_u}$ and item embedding look-up table $V \in R^{m \times d_i}$, where $n$ is the number of users, $m$ is the number of items and $d_u$, $d_i$ are the number of hidden units for user and item respectively. We also specify learnable positional encoding table $P \in R^{T \times d}$, where $d = d_u + d_i$.
So each input sequence 
$s_i \in R^T$ will be represented by the following embedding: 
\begin{equation}\label{eq:emb}
    E = \begin{bmatrix}
    [v_{j_{i1}}\text{; } u_i] + p_1   \\
    [v_{j_{i2}}\text{; } u_i] + p_2 \\
    \vdots \\
    [v_{j_{iT}}\text{; } u_i] + p_T
    \end{bmatrix} \in R^{T \times d},
\end{equation} 
where $[v_{j_{it}}; u_i]$ represents concatenating item embedding $v_{j_{it}} \in R^{d_i}$ and user embedding $u_i \in R^{d_u}$ into embedding $E_t \in R^d$ for time $t$. Note that the main difference between our model and \cite{kang2018self} is that we introduce the user embeddings $u_i$, making our model personalized.

\subsubsection{Self-Attention Layer}
Self-attention layers defined as:
\begin{equation}
    S = \text{SA}(E) = \text{Attention}\left(E W^{(Q)}\text{, } E W^{(H)}\text{, } E W^{(V)}\right),
\end{equation} 
where $W^{(Q)}, W^{(H)}, W^{(V)} \in R^{d \times d}$ and  
\begin{equation}
     \text{Attention}(Q, H, V) = \text{softmax}\left(\frac{Q H^T}{\sqrt{d}} \right)\cdot V.
\end{equation}
    
    The attention layer we actually used is a masked one because we only want attention from the future to the past,
    not the opposite direction. Therefore, all links between $Q_i, H_j$ for $j > i$ are forbidden. We find using bidirectional attention would lead to significantly worse performance.
    
\subsubsection{Pointwise Feed-Forward layer}
After feeding embeddings into the self-attention layer, we want to add non-linearity to the resulting $S\in R^{n \times d}$ for each sequence data. Therefore, we add a pointwise feed-forward layer after the self-attention layer, consisting of two fully connected layers:
\begin{equation}
    F = \text{FC}(S) = \text{Relu}(S W + b) \cdot \tilde{W} + \tilde{b},
\end{equation} 
where $W$, $\tilde{W} \in R^{d \times d}$ are the weight matrices and $b$, $\tilde{b} \in R^{d}$ are the bias terms.

\subsubsection{Self-Attention Blocks}
We combine self-attention layer and pointwise feed-forward layer to form a self-attention (SA) block.
One block consists of one self-attention layer and two fully connected layers. We can stack blocks by feeding the output of first block $F^{(1)}$ as the input of the second block, i.e. 
\begin{equation}
    S^{(2)} = \text{SA}(F^{(1)}).
\end{equation}
 We use $B$ to denote the number of attention blocks. 

\begin{figure}[ht]
\vskip -0.1in
\begin{center}
\centerline{\includegraphics[width=0.8\columnwidth]{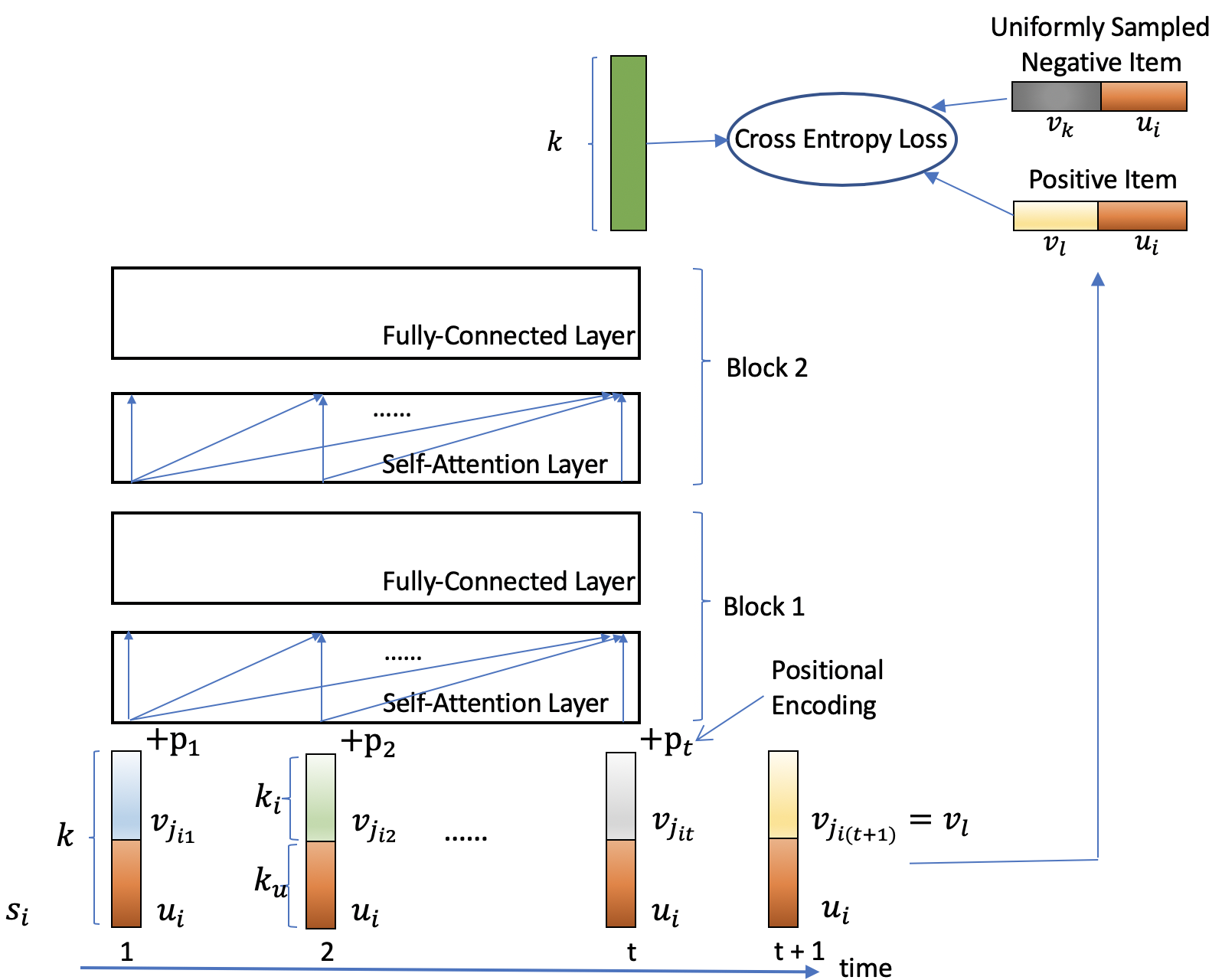}}  
\end{center}
\vskip -0.2in
\caption{Illustration of our proposed SSE-PT model}
\label{fig:SSE-PT}
\vspace{-10pt}\end{figure}

\subsubsection{Prediction Layer}
As to the prediction layer, the predicted probability of user $i$ at time $t$ rated item $l$ is: 
\vspace{-5pt}\begin{equation}
    p_{itl} = \sigma(r_{itl}) 
\end{equation} where
$\sigma$ is the sigmoid function and $r_{itl}$ is the predicted score of item $l$ by user $l$ at time point $t$, defined as:
\begin{equation}\label{eq:pred}
        r_{itl} = F_{t-1}^{B} \cdot [v_l\text{; } u_i].
\end{equation}
Although we can use another set of user and item embedding look-up tables for the $u_i$ and $v_l$, we find it better to use the same set of embedding look-up tables $U, V$ as in the embedding layer. To distinguish the $u_i$ and $v_l$ in \eqref{eq:pred} from $u_i, v_j$ in \eqref{eq:emb}, we call embeddings in \eqref{eq:pred} output embeddings and those in \eqref{eq:emb} input embeddings.

There are multiple ways to define the loss of our model, previously a popular loss is the BPR loss \cite{rendle2009bpr, hidasi2018recurrent}:
\begin{equation}\label{eq:bprloss}
    \sum\nolimits_{i}\sum\nolimits_{t=1}^{T-1} -\sum_{k \in \Omega} \log \big[ \sigma (r_{itl} - r_{itk})\big],
\end{equation} where $\sigma$ is the sigmoid function, $r_{itl}$ is the predicted score of the positive item $l$ at time point $t$ for user $i$, $r_{itk}$ is the predicted score of the negative item, and set of negative items is defined as $\Omega = \{1 \leq j \leq m : j \neq j_{it}, \forall 1 \leq t \leq T \}$. At time point $t$, the positive item index is $l = j_{i(t+1)}$ in \eqref{eq:input}, and negative item index $k$ satisfies $k \in \Omega$.

We find the BPR loss does not perform as well as the binary cross entropy loss in practice.
The binary cross entropy loss between predicted probability for the positive item $l = j_{i (t+1)}$ and one uniformly sampled negative item $k \in \Omega$ is given as
$-[\log (p_{itl}) + \log(1 - p_{itk})]$. 
Summing over $s_i$ and $t$, we obtain the objective function that we want to minimize is:
\begin{equation}
    \sum\nolimits_{i}\sum\nolimits_{t=1}^{T-1} \sum_{k\in \Omega}-\big[\log (p_{itl}) +  \log(1 - p_{itk})\big].
\end{equation}
At the inference time, top-$K$ recommendations for user $i$ at time $t$ can be made by sorting scores  $r_{itl}$ for all items $\ell$ and recommending the first $K$ items in the sorted list.

\subsection{Personalized Transformer Regularization Techniques}
 \subsubsection{Layer Normalization}
 Layer normalization \cite{ba2016layer} normalizes neurons within a layer. Previous studies \cite{ba2016layer} show it is more effective than batch normalization for training recurrent neural networks (RNNs).
One alternative is the batch normalization \cite{ioffe2015batch} but we find it does not work as well as the layer normalization in practice even for a reasonable large batch size of 128. Therefore, our SSE-PT model adopts layer normalization.  

\subsubsection{Residual Connections}
Residual connections are firstly proposed in ResNet for image classification problems \cite{he2016deep}. 
Recent research finds that residual connections can help training very deep neural networks even if they are not convolutional neural networks \cite{vaswani2017attention}.  
Using residual connections allows us to train very deep neural networks here. For example, the best performing model for Movielens10M dataset in Table~\ref{tb:block} is the SSE-PT with 6 attention blocks, in which $1 + 6 * 3 + 1 = 20$ layers are trained end-to-end.

\subsubsection{Weight Decay}
Weight decay \cite{krogh1992simple}, also known as $l_2$ regularization \cite{hoerl1970ridge}, is applied to all embeddings, including both user and item embeddings.

\subsubsection{Dropout}
Dropout \cite{srivastava2014dropout} is applied to the embedding layer $E$, self-attention layer and pointwise feed-forward layer by stochastically dropping some percentage of hidden units to prevent co-adaption of neurons. Dropout has been shown to be an effective way of regularizing deep learning models.

\subsubsection{Stochastic Shared Embeddings}
Unlike previous SASRec model \cite{kang2018self}, we use one more regularization technique in our SSE-PT model specifically for embedding layer in addition to the ones listed earlier: the Stochastic Shared Embeddings (SSE) \cite{wu2019stochastic}. The reason that we want to use this additional regularization technique is that we find the existing well-known regularization techniques like layer normalization, dropout and weight decay cannot prevent the model from over-fitting badly after introducing user embeddings. We apply this new regularization technique  SSE-SE to our SSE-PT model, and we find it
makes possible training this personalized model with $O(n d_u)$ more parameters.

The main idea of SSE is to stochastically replace embeddings with another embedding during SGD, which has the effect of regularizing the embedding layers.
Specifically, SSE-SE replaces one embedding with another embedding stochastically with probability $p$, which is called SSE probability in \cite{wu2019stochastic}. There are 3 different places in our model that SSE-SE can be applied. We can apply SSE-SE to input/output user embeddings, input item embeddings, and output item embeddings with probabilities $p_u$, $p_i$ and $p_y$ respectively. Note that input user embedding and output user embedding are always replaced at the same time with SSE probability $p_u$. Empirically, we find that SSE-SE to user embeddings and output item embeddings always helps, but SSE-SE to input item embeddings is only useful when the average sequence length is large, e.g. more than 100 in Movielens1M and Movielens10M datasets.

In summary, layer normalization and dropout are used in all layers except prediction layer. Residual connections are used in both self-attention layer and pointwise feed-forward layer. SSE-SE is used in embedding layer and prediction layer.

\subsection{Handling Long Sequences: SSE-PT++}
 To handle extremely long sequences, a slight modification can be made on the way how input sequences $s_i$'s are fed into the SSE-PT neural networks. We call the enhanced model SSE-PT++ to distinguish it from the standard SSE-PT model, which cannot handle sequences longer than $T$.   
 
 Sometimes, we want to make use of extremely long sequences, $s_i=(j_{i1}, j_{i2}, \dots, j_{it}) \text{ for } 1 \leq i \leq n$, where $t > T$. However, our SSE-PT model can only handle sequences of maximum length of $T$. 
 The simplest way is to sample starting index $1 \leq v \leq t$ uniformly and use $s_i=(j_{iv}, j_{i(v+1)}, \dots, j_{iz})$, where $z = \min (t, v + T)$. Although sampling the starting index uniformly from $[1, t]$ can accommodate long sequences of length $t > T$, this does not work very well in practice. Uniformly sampling does not take into account the importance of recent items in a long sequence. To solve this dilemma, we introduce an additional hyper-parameter $p_s$ which we call {\it sampling probability}. It implies that with probability $p_s$, we sample the starting index $v$ uniformly from $[1, t - T]$ and use sequence $s_i=(j_{iv}, j_{i(v+1)}, \dots, j_{i(v+T-1)})$ as input. With probability $1 - p_s$, we will simply use the recent $T$ items  $(j_{i(t-T+1)}, \dots, j_{it})$ as input. If the sequence $s_i$ is already shorter than $T$ then we simply set $p_s = 0$ for user $i$.
 
 Our proposed SSE-PT++ model can work almost as well as SSE-PT with a much smaller $T$. One can see in Table~\ref{tb:dl-ml2} with $T = 100$ SSE-PT++ can perform almost as well as SSE-PT. The time complexity of the SSE-PT model is of order $O(T^2 d + T d^2)$. Therefore, reducing $T$ by one half would lead to a theoretically 4x speed-up in terms of the training and inference speeds. As to space complexity, both SSE-PT and SSE-PT++ are of order $O(nd_u + md_i + Td + d^2)$. When the number of users and items scales up, Tensorflow will automatically store the user and item embedding look-up tables in RAM instead of GPU memory.


\begin{figure*}[ht]
\begin{center}
\centerline{\includegraphics[width=1\columnwidth]{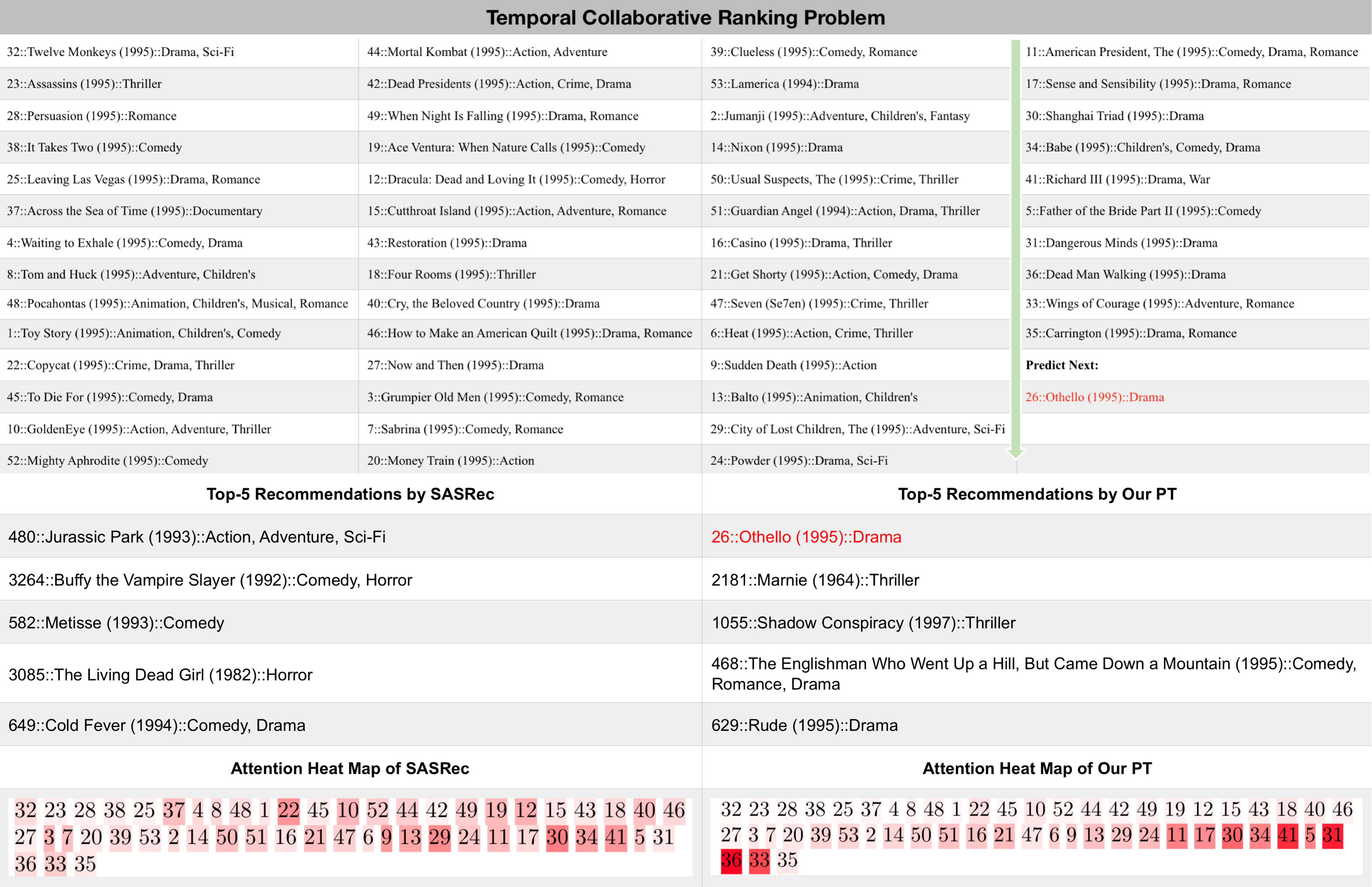}}
\end{center}
\caption{Illustration of how SASRec (Left) and SSE-PT (Right) differs on utilizing the Engagement History of A Random User in Movielens1M Dataset.}
\label{fig:example}
\end{figure*}

\begin{table}[ht]
\centering
\vskip -0.15in
\caption{Description of Datasets Used in Evaluations.}
\vskip -0.1in
\label{tb:datasets}
\begin{center}
\begin{small}
\begin{sc}
\resizebox{0.6\textwidth}{!}{
\begin{tabular}{ccccc}
\toprule
 dataset & \#users & \#items & avg sequence len & max sequence len   \\
\midrule
\multirow{1}{*}  Beauty   &  52,024  & 57,289   & 7.6 & 291 \\
games   &  31,013 &  23,715   & 7.3 & 858 \\
steam & 334,730 &  13,047 &   11.0 & 1,229  \\
ml-1m  &  6,040  &  3,416 & 163.5  & 2,275  \\
ml-10m  & 69,878  & 65,133 &  141.1 & 7,357  \\
\bottomrule
\end{tabular}
}
\end{sc}
\end{small}
\end{center}
\vskip -0.1in
\end{table}

\section{Experiments}
In this section, we compare our proposed algorithms, Personalized Transformer (SSE-PT) and SSE-PT++, with other state-of-the-art algorithms on real-world datasets. We implement our codes in Tensorflow and conduct all our experiments on 

a server with 40-core Intel Xeon E5-2630 v4 @ 2.20GHz CPU, 256G RAM and Nvidia GTX 1080 GPUs.

\subsection{Datasets}
We use 5 datasets, the first 4 of them have exactly the same train/dev/test splits as in \cite{kang2018self}:
\begin{itemize}
    \item Beauty category from Amazon product review datasets. \footnote{\url{http://jmcauley.ucsd.edu/data/amazon/}} 
    \item Games category from the same source.
    \item Steam dataset introduced in \cite{kang2018self}. It contains reviews crawled from a large video game distribution platform. 
    \item Movielens1M dataset \cite{harper2016movielens}, a widely used benchmark datasets containing one million user movie ratings.
    \item Movielens10M dataset with ten million user ratings.
\end{itemize}
Detailed dataset statistics are given in Table~\ref{tb:datasets}. One can easily see that the first 3 datasets have very short sequences while the last 2 datasets have very long sequences.

\subsection{Evaluation Metrics}
The evaluation metrics we use are standard ranking metrics, namely NDCG and Recall for top recommendations:
\begin{itemize}
    \item NDCG$@K$: defined as:
        \begin{equation}\label{eq:ndcg}
            \text{NDCG}@K = \frac{1}{n} \sum_{i = 1}^{n} \frac{\text{DCG}@K(i, \Pi_i)}{\text{DCG}@K(i, \Pi_i^*)}, 
        \end{equation} where $i$ represents $i$-th user and
        \begin{equation} 
        \text{DCG}@K(i, \Pi_i)= \sum_{l = 1}^{K} \frac{2^{R_{i\Pi_{il}}} - 1}{\log_2(l + 1)}. 
        \end{equation}
        In the DCG definition, $\Pi_{il}$ represents the index of the $l$-th ranked item for user $i$ in test data based on the learned score matrix $X$. $R$ is the rating matrix and $R_{ij}$ is the rating given to item $j$ by user $i$. $\Pi_i^*$ is the ordering provided by the ground truth rating.

    \item Recall@$K$: defined as a fraction of positive items retrieved by the top $K$ recommendations the model makes: 
        \begin{equation}\label{eq:recall}
            \text{Recall}@K = \frac{\sum_{i=1}^{n} \mathbbm{1} \{  \exists 1 \leq l \leq K : R_{i\Pi_{il}} = 1 \}} {n},
        \end{equation}
        here we already assume there is only a single positive item that user will engage next and the indicator function $\mathbbm{1} \{  \exists 1 \leq l \leq k : R_{i\Pi_{il}} = 1 \}$ is defined to indicate whether the positive item falls into the top $K$ position in our obtained ranked list using scores predicted in \eqref{eq:pred}. 
\end{itemize}

\begin{figure*}[t]
\centering
\begin{tabular}{cc}
\includegraphics[width=0.5\linewidth]{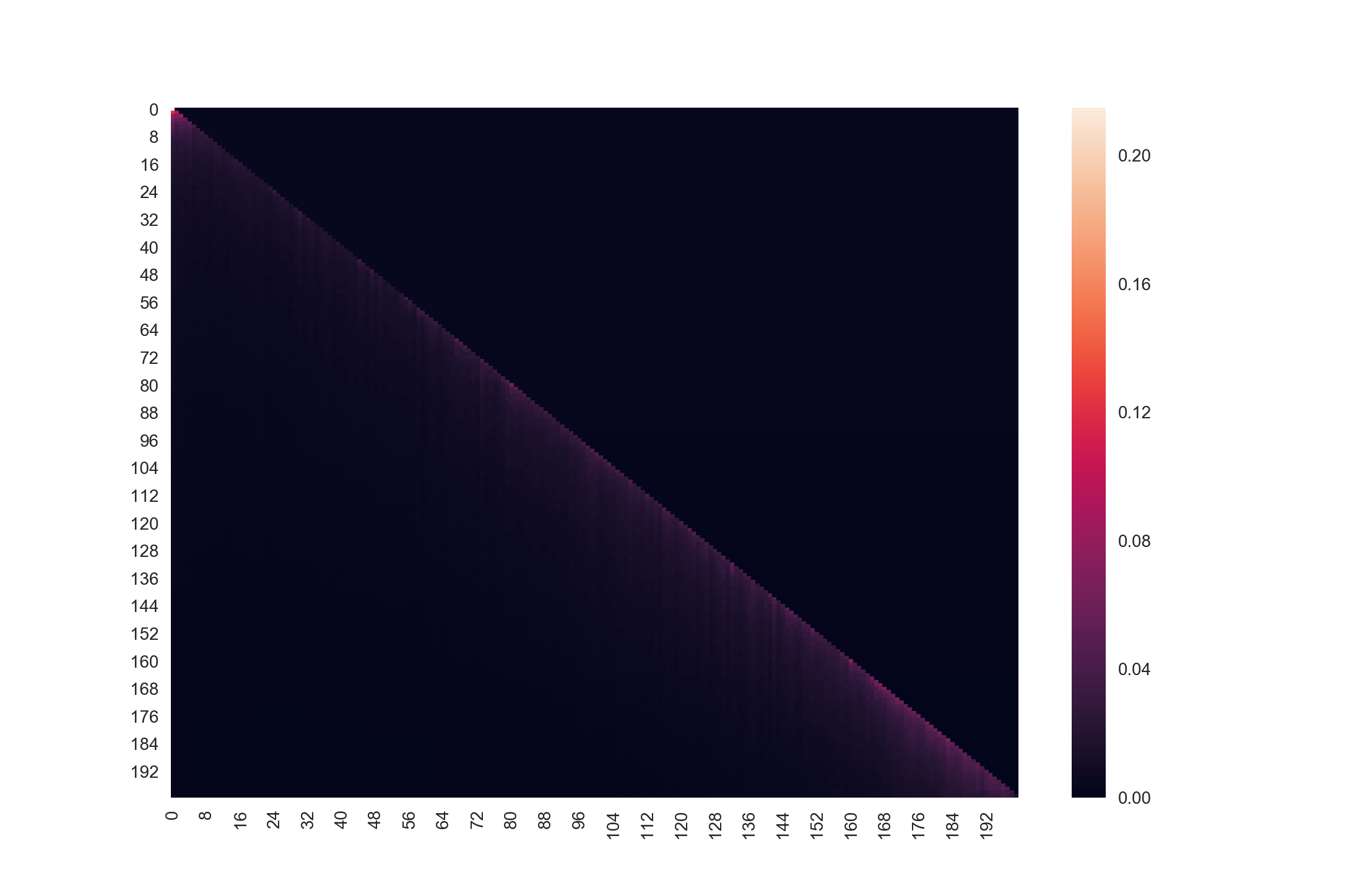} &
\includegraphics[width=0.48\linewidth]{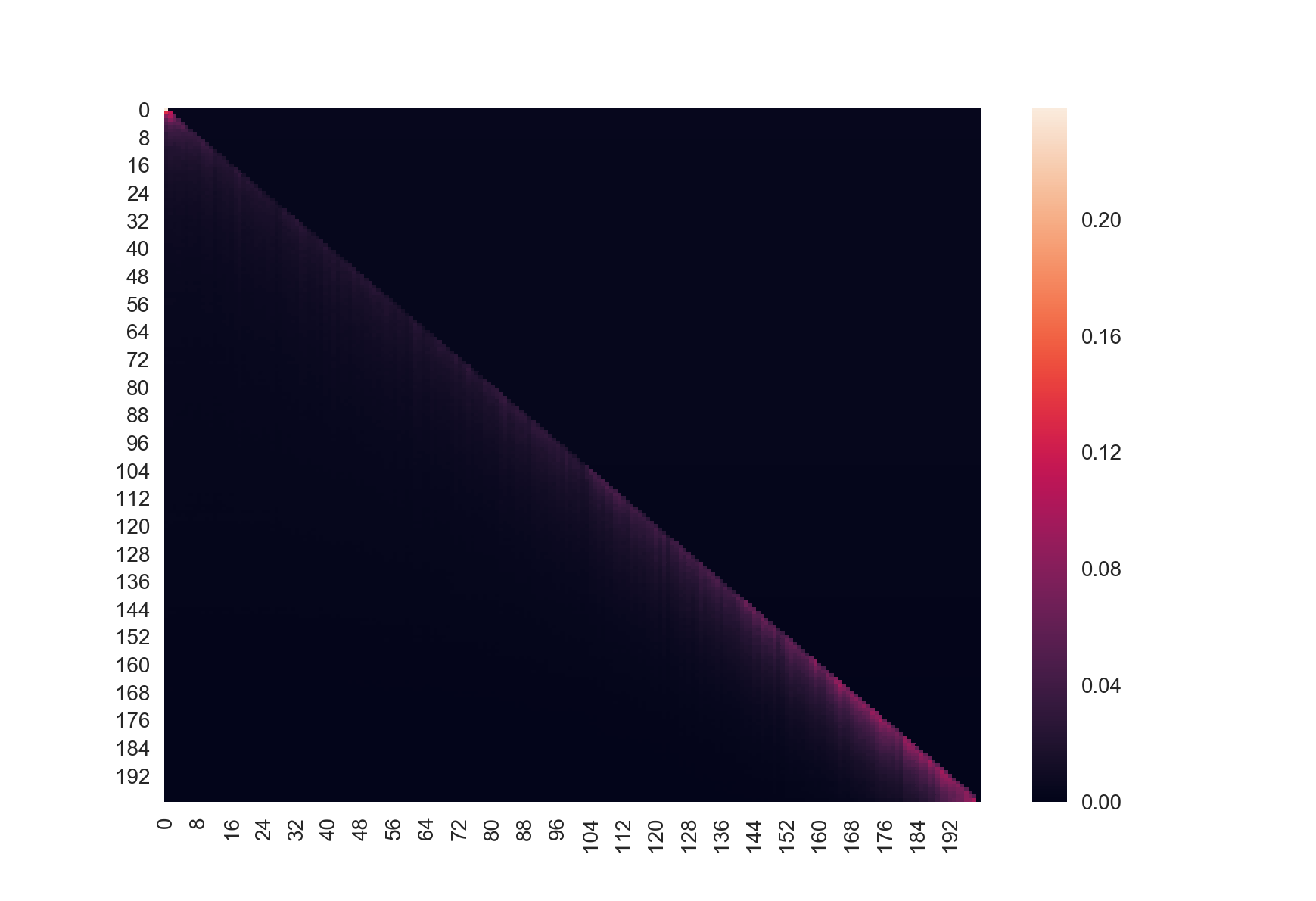} 
\end{tabular}
\vspace{-10pt}
\caption{Compare Attention Maps for Layer-1 between current state-of-the-art SASRec (Left) and our proposed SSE-PT (Right). }
\label{fig:attention}
\vspace{-10pt}
\end{figure*}

In the temporal collaborative ranking setting, at time point $t$, the rating matrix $R$ can be formed in two ways. One is that we include all ratings after $t$, the other is to include only ratings at time point $t + 1$. We use the latter, which is the same setting as \cite{kang2018self}. 
For a large dataset with numerous users and items, the evaluation procedure would be slow because \eqref{eq:ndcg} would require computing the ranking of all items based on their predicted scores for every single user. As a means of speed-up evaluations, we sample a fixed number $C$ of negative candidates while always keeping the positive item that we know the user will engage next. This way, both $R_{ij}$ and $\Pi_i$ will be narrowed down to a small set of item candidates, and prediction scores will only be computed for those items through a single forward pass of the neural network.

Ideally, we want both NDCG and Recall to be exactly 1, because NDCG@$K = 1$ means the positive item is always put on the top-$1$ position of the top-$K$ ranking list, and Recall@$K = 1$ means the positive item is always contained by the top-$K$ recommendations the model makes. In our evaluation procedures, a larger $C$ or a smaller $K$ makes the recommendation problem harder because it implies the candidate pool is larger and higher ranking quality is desired.

\begin{table*}[ht]
\centering
\caption{Comparing various state-of-the-art temporal collaborative ranking algorithms on various datasets. Note that (F) to (K) are deep learning methods, we use underline to highlight best result for non-deep-learning methods and bold fonts to highlight best overall result.}
\label{tb:dl-combined}
\begin{center}
\begin{small}
\begin{sc}
\resizebox{\textwidth}{!}{
\begin{tabular}{lccccccccccccr}
\toprule
& &(A) & (B) & (C) & (D) & (E) & (F) & (G) & (H) & (I) & (J)  & (K)& \% GAIN OVER \space\space\\[-1ex]
\raisebox{1.5ex}{DATASET} & \raisebox{1.5ex}{METRIC}   
&  POPREC & BPR & FMC & FPMC & TRANSREC  &GRU4REC & STAMP & GRU4REC+ & CASER & SASREC & SSE-PT & (A)-(E) \space\space (F)-(J) \\[-0.2ex]

\midrule

& Recall@$10$ & 0.4003 & 0.3775 & 0.3771 & 0.4310 & \underline{0.4607}  & 0.2125 & 0.3207 & 0.3949 & 0.4264 & 0.4837  &\textbf{0.5028}  & 13.0 \space\space \space \space \space \space 3.6 \space  \space\space\\[-1ex]
\raisebox{1.1ex}{BEAUTY} & NDCG@$10$ & 0.2277 & 0.2183 & 0.2477 & 0.2891 & \underline{0.3020} & 0.1203 & 0.1801 & 0.2556 & 0.2547 & 0.3220 & \textbf{0.3370} & 11.6 \space\space \space \space \space \space 4.7\space \space \space\space \\[1.0ex] 

& Recall@$10$ & 0.4724 & 0.4853 & 0.6358 & 0.6802 & \underline{0.6838}& 0.2938  & 0.4358 & 0.6599 & 0.5282& 0.7434 & \textbf{0.7757} & 13.4 \space\space \space \space \space \space 4.3 \space  \space\space\\[-1ex]
\raisebox{1.1ex}{GAMES} & NDCG@$10$ & 0.2779 & 0.2875 & 0.4456 & \underline{0.4680} & 0.4557 & 0.1837  & 0.2615& 0.4759 & 0.3214 & 0.5401 & \textbf{0.5660} & 20.3 \space\space \space \space \space \space 4.8\space \space \space\space\\[1.0ex] 

& Recall@$10$ & 0.7172 & 0.7061 & \underline{0.7731} & 0.7710 & 0.7624 & 0.4190  & 0.6982 & 0.8018 & 0.7874 & 0.8732 & \textbf{0.8772} & 13.5 \space \space \space \space \space\space 0.4 \space \space \space\\[-1ex]
\raisebox{1.1ex}{STEAM} & NDCG@$10$ & 0.4535 & 0.4436 & \underline{0.5193} & 0.5011 & 0.4852 &0.2691 & 0.4296 & 0.5595 & 0.5381 & 0.6293 & \textbf{0.6378} & 22.8 \space \space \space \space \space \space 1.4\space \space \space\space \\[1.0ex]

& Recall@$10$ & 0.4329 & 0.5781 & 0.6986 & \underline{0.7599} & 0.6413 & 0.5581 & 0.7255 & 0.7501 & 0.7886 & 0.8233 & \textbf{0.8358} & 10.0 \space \space \space \space \space\space 1.5 \space \space \space\\[-1ex]
\raisebox{1.1ex}{ML-$1$M} & NDCG@$10$ & 0.2377 & 0.3287& 0.4676 & \underline{0.5176} & 0.3969 &0.3381  & 0.4813 & 0.5513 & 0.5538 & 0.5936 & \textbf{0.6191} & 19.6 \space \space \space \space \space \space4.3\space \space \space\space \\[-0.5ex]

\bottomrule
\end{tabular}
}

\end{sc}
\end{small}
\end{center}
\vskip 0.1in
\end{table*}

\begin{table}[ht]
\centering
\vskip -0.05in
\caption{Comparing SASRec, SSE-PT and SSE-PT++ on Movielens1M Dataset while varying dimension of embeddings.}
\vskip -0.1in
\label{tb:dl-ml}
\vskip -0.15in
\begin{center}
\begin{small}
\begin{sc}
\resizebox{0.6\textwidth}{!}{
\begin{tabular}{lccccr}
\toprule
 Methods & NDCG@$10$ & Recall@$10$ & user dim & item dim \\
\midrule
\multirow{1}{*}  SASREC   &  0.5936  & 0.8233   & N/A & 50 \\
 SASREC   &  0.5919 & 0.8202   & N/A & 100 \\
 SSE-PT & 0.6191 & 0.8358 &  50 & 50  \\
 SSE-PT & 0.6281 & 0.8341 &  50 & 100  \\
 SSE-PT++ & \bfseries{0.6292} & \bfseries{0.8389} &  50 & 100  \\
\bottomrule
\end{tabular}
}
\end{sc}
\end{small}
\end{center}
\end{table}

\begin{table}[ht]
\centering
\vskip -0.05in
\caption{Comparing SASRec, SSE-PT and SSE-PT++ on Movielens1M Dataset while varying the maximum length allowed.}
\vskip -0.1in
\label{tb:dl-ml2}

\begin{center}
\begin{small}
\begin{sc}
\resizebox{0.6\textwidth}{!}{
\begin{tabular}{lcccccc}
\toprule
 Methods & NDCG@$10$ & Recall@$10$ & Max Len & user dim & item dim \\
\midrule
 SASREC   &  0.5919 & 0.8202  &200 & N/A & 100 \\
  SSE-PT & \bfseries{0.6281} & \bfseries{0.8341} &200 &  50 & 100  \\
  \midrule
 SASREC   &  0.5769  & 0.8045   & 100 & N/A & 100 \\
 SSE-PT & 0.6142 & 0.8212 & 100 &  50 & 100  \\
 SSE-PT++ & \bfseries{0.6186} & \bfseries{0.8318} & 100 &  50 & 100  \\
\bottomrule
\end{tabular}
}
\end{sc}
\end{small}
\end{center}
\vskip -0.15in
\end{table}

\subsection{Baselines}
We include 5 non-deep-learning and 5 deep-learning algorithms in our comparisons:
\subsubsection{Non-deep-learning Baselines}
\begin{itemize}
    \item PopRec: ranking items according to their popularity. 
    \item BPR: Bayesian personalized ranking for implicit feedback setting \cite{rendle2009bpr}. It is a low-rank matrix factorization model with a pairwise loss function.
    But it does not utilize the temporal information. Therefore, it serves as a strong baseline for non-temporal methods.
     \item FMC: Factorized Markov Chains: a first-order Markov Chain method, in which 
     predictions are made only based on previously engaged item.  
     \item PFMC: a personalized Markov chain model \cite{rendle2010factorizing} that combines matrix factorization and first-order Markov Chain to take advantage of both users' latent long-term preferences as well as short-term item transitions.
    \item TransRec: a first-order sequential recommendation method \cite{he2017translation} in which items are embedded into a transition space and users are modelled as translation vectors operating on item sequences.
\end{itemize}
SQL-Rank \cite{wu2018sql} and item-based recommendations \cite{sarwar2001item}  are omitted
because the former is similar to BPR \cite{rendle2009bpr} except using the listwise loss function instead of the pairwise loss function and the latter has been shown inferior to TransRec \cite{he2017translation}.

\begin{table*}[t]
\vskip -0.1in
  \caption{Comparing our SSE-PT, SSE-PT++ with SASRec on Movielen1M dataset. We use number of negatives $C = 100$, dropout probability of $0.2$ and learning rate of $1e^{-3}$ for all experiments while varying others. $p_u, p_i, p_u$ are SSE probabilities for user embedding, input item embedding and output item embedding respectively.}
  \vskip -0.01in
  \label{tb:ml1m}
  \centering
 \resizebox{1\columnwidth}{!}{
  \begin{tabular}{lccccccccc}
    \toprule
    & \multicolumn{2}{c}{Movielens1m }  & \multicolumn{2}{c}{Dimensions} & Number of Blocks & Sampling Probability & \multicolumn{3}{c}{SSE-SE Parameters}             \\
    \cmidrule(r){2-3} \cmidrule(r){4-5} \cmidrule(r){6-6} \cmidrule(r){7-7}  \cmidrule(r){8-10}
    Model     & NDCG$@10$ & Recall$@10$ & $d_u$ & $d_i$ & $b$ & $p_s$ & $p_u$ & $p_i$  & $p_y$ \\
    \midrule
    SASRec            & 0.5961 & 0.8195& - & 50 & 2 & - & -  & - & - \\
    SASRec   & 0.5941 & 0.8182 & - & 100 & 2  & - & - & -  & -\\
    SASRec   & \bfseries{0.5996} & \bfseries{0.8272} & - & 100 & 6 & - & - & -  & -\\
     \midrule
    SSE-PT  & 0.6101 & 0.8343 & 50 &  50 & 2 & -& 0.92 & 0.1  & 0 \\
    SSE-PT  & 0.6164 & 0.8336 & 50 &  50 & 2 & -& 0.92 & 0  & 0.1 \\
    SSE-PT  & 0.5832 & 0.8091 & 50 &  50 & 2 & -& 0 & 0.1  & 0.1 \\
    SSE-PT  & \bfseries{0.6174}  & \bfseries{0.8351} & 50 &  50 & 2& - & 0.92 & 0.1  & 0.1 \\
    \midrule
     SSE-PT  & 0.5949  & 0.8205 & 75 &  25 & 2 & -& 0.92 & 0.1  & 0.1 \\
     
   
   SSE-PT  & \bfseries{0.6214}  & \bfseries{0.8359} & 25 &  75 & 2& - & 0.92 & 0.1  & 0.1 \\
    \midrule
    SSE-PT  & 0.6281 & 0.8341 & 50 & 100 & 2 & -& 0.92 & 0.1  & 0.1 \\
    SSE-PT++ & \bfseries{0.6292} & \bfseries{0.8389} &  50 & 100 & 2& 0.3 & 0.92 & 0.1  & 0.1 \\
    \bottomrule
  \end{tabular}
}
\end{table*}

\begin{table}[ht]
\centering
\caption{Comparing Different Regularization Techniques for SSE-PT on Movielen1M Dataset. NO REG stands for no regularization.
PS stands for parameter sharing across all users while PS(AGE) means PS is used within each age group. 
SASRec is added to last row after all SSE-PT results as a baseline.}
\label{tb:reg}
\begin{center}
\begin{small}
\begin{sc}
\resizebox{0.7\textwidth}{!}{
\begin{tabular}{lcccccr}
\toprule
Regularization & NDCG@$5$ & $\%$ GAIN & Recall@$5$ & $\%$ GAIN \\
\midrule
NO REG (BASELINE)  & 0.4855 & - & 0.6500 &  -   \\
PS            & 0.5065 & 4.3 & 0.6656 & 2.4  \\
PS (JOB)       & 0.4938 & 1.7 & 0.6570 & 1.1  \\
PS (GENDER)    & 0.5110 & 5.3 & 0.6672 & 2.6  \\
PS (AGE)       & 0.5133 & 5.7 & 0.6743 & 3.7  \\
$l_2$            & 0.5149 & 6.0 & 0.6786 & 4.4  \\
DROPOUT       & 0.5165 & 6.4 & 0.6823 & 5.0  \\
$l_2$  + DROPOUT & 0.5293 & 9.0  & 0.6921 & 6.5  \\
 SSE-SE            & 0.5393 & 11.1 & 0.6977 & 7.3  \\
$l_2$ + SSE-SE + DROPOUT &  \bfseries{0.5870} & \textbf{20.9} & \textbf{0.7442} & \textbf{14.5}  \\
\midrule
SASRec ($l_2$ + DROPOUT)& 0.5601 & & 0.7164 &  \\

\bottomrule
\end{tabular}
}
\end{sc}
\end{small}
\end{center}
\end{table}

\begin{table*}[t]
  \caption{Comparing our SSE-PT with SASRec on Movielens10M dataset. Unlike Table~\ref{tb:ml1m}, we use the number of negatives $C = 500$ instead of $100$ as $C = 100$ is too easy for this dataset and it gets too difficult to tell the differences between different methods.}
  \label{tb:ml10m}
  \centering
 \resizebox{0.9\columnwidth}{!}{
  \begin{tabular}{ccccccccc}
    \toprule
    & \multicolumn{2}{c}{Movielens1m }  & \multicolumn{2}{c}{Dimensions} & Number of Blocks  & \multicolumn{3}{c}{SSE-SE Parameters}             \\
    \cmidrule(r){2-3} \cmidrule(r){4-5} \cmidrule(r){6-6}  \cmidrule(r){7-9}
    Model     & NDCG$@10$ & Hit Ratio$@10$ & $d_u$ & $d_i$ & $b$ & $p_u$ & $p_i$  & $p_y$ \\
    \midrule
    SASRec   & 0.7268 & 0.9429 & - & 50 & 2  & - & -  & -\\
    SASRec   & 0.7413 & 0.9474 & - & 100 & 2  & - & -  & -\\
    \midrule
   SSE-PT  & 0.7199 & 0.9331 & 50 &  100 & 2 & PS & 0.01  & 0.01 \\
   SSE-PT  & 0.7169 & 0.9296 & 50 &  100 & 2 & 0.0 & 0.01  & 0.01 \\
    SSE-PT  & 0.7398 & 0.9418 & 50 &  100 & 2 & 0.2 & 0.01  & 0.01 \\
    SSE-PT  & 0.7500 & 0.9500 & 50 &  100 & 2 & 0.4 & 0.01  & 0.01 \\
    SSE-PT  & 0.7484 & 0.9480 & 50 &  100 & 2 & 0.6 & 0.01  & 0.01 \\
    SSE-PT  & \bfseries{0.7529} & 0.9485 & 50 &  100 & 2 & 0.8 & 0.01  & 0.01 \\
    SSE-PT  & 0.7503 &  \bfseries{0.9505} & 50 &  100 & 2 & 1.0 & 0.01  & 0.01 \\
    \bottomrule
  \end{tabular}
}
\vspace{-10pt}
\end{table*}

\subsubsection{Deep-learning baselines}
\begin{itemize}
    \item GRU4Rec: the first RNN-based method proposed for the session-based recommendation problem \cite{hidasi2015session}. It utilizes the GRU structures \cite{chung2014empirical} initially proposed for speech modelling.
    \item GRU4Rec$^+$: follow-up work of GRU4Rec by the same authors: the model has a very similar architecture to GRU4Rec but has a more complicated loss function \cite{hidasi2018recurrent}.
    \item Caser: a CNN-based method \cite{tang2018personalized} which embeds a sequence of recent items in both time and latent spaces forming an `image' before learning local features through horizontal and vertical convolutional filters. In \cite{tang2018personalized}, user embeddings are included in the prediction layer only. On the contrast, in our Personalized Transformer, user embeddings are also introduced in the lowest embedding layer so they can play an important role in self-attention mechanisms as well as in prediction stages.
    \item STAMP: a session-based recommendation algorithm \cite{liu2018stamp} using attention mechanism. \cite{liu2018stamp} only uses fully connected layers with one attention block that is not self-attentive. 
    \item SASRec: a self-attentive sequential recommendation method \cite{kang2018self} motivated by Transformer in NLP \cite{vaswani2017attention}. Unlike our method SSE-PT, SASRec does not incorporate user embedding and therefore is not a personalized method. SASRec paper \cite{kang2018self} also does not utilize SSE \cite{wu2019stochastic} for further regularization: only dropout and weight decay are used.
\end{itemize}

\subsection{Comparison Results}
We use the same datasets as in \cite{kang2018self} and follow the same procedure in the paper: use last items for each user as test data, second-to-last as validation data and the rest as training data.  We implemented our method in Tensorflow and solve it with Adam Optimizer \cite{kingma2014adam} with a learning rate of $0.001$, momentum exponential decay rates $\beta_1 = 0.9, \beta_2 = 0.98$ and a batch size of $128$. In Table 3, since we use the same data, the performance of previous methods except STAMP have been reported in \cite{kang2018self}. We tune the dropout rate, and SSE probabilities $p_u, p_i, p_y$ for input user/item embeddings and output embeddings on validation sets 
and report the best NDCG and Recall for top-$K$ recommendations on test sets. As mentioned before, we sampled $C$ negative items to speed up the evaluation. 

For a fair comparison,
we restrict all algorithms to use up to 50 hidden units for item embeddings.
For the SSE-PT and SASRec models, we use the same number of attention blocks of 2 and set the maximum length $T = 200$ for Movielens 1M dataset and $T = 50$ for other datasets.
We use top-$K$ with $K = 10$ and number of negatives $C = 100$ in evaluation procedure.
One can easily see in Table~\ref{tb:dl-combined} that our proposed SSE-PT has the best performance over all previous methods on all four datasets we consider. On most datasets, our SSE-PT improves NDCG by more than 4\% when compared with SASRec \cite{kang2018self} and more than 20\% when compared to non-deep-learning methods.

When we relax the constraints, we find that an increase in the number of attention blocks and hidden units  would allow our SSE-PT model to perform even better than in Table~\ref{tb:dl-combined}. In Table~\ref{tb:ml1m}, when we increase item embedding dimension $d_i$ from 50 to 100, our SSE-PT achieves 0.6281 for NDCG@10 and SSE-PT++ achieves even higher 0.9292 while that of SASRec drops to 0.5919 from 0.5936. 

To show the effectiveness of SSE-PT++, we decrease max length allowed from 200 to 100, we find in Table~\ref{tb:dl-ml2} that SSE-PT++ suffer the least with NDCG@10 dropping to 0.6186 from 0.6281 and Recall@10 dropping to 0.8318 from 0.8341. The one that suffers the most is the SASRec: its NDCG@10 drops to 0.5769 from 0.5919 and Recall@10 drops to 0.8045 from 0.8202.

We vary the tuning parameters we use in Table~\ref{tb:ml1m} including user/item embedding dimensions, the number of attention blocks, SSE probabilities for SSE-PT and the sampling probability for SSE-PT++. It is obvious from Table~\ref{tb:ml1m} and Table~\ref{tb:ml10m} that these hyper-parameters play an important role in terms of the final prediction performances. For our SSE-PT model, a larger item dimension helps improve the recommendation but it is not the case for baseline SASRec. Also, using SSE-SE in all three places achieves best recommendation performances for Movielens1M dataset in Table~\ref{tb:ml1m}. One can easily see from Table~\ref{tb:ml10m}, using SSE-SE towards input user embeddings is crucial again to ensure a properly regularized model. SSE-SE, together with dropout and weight decay, is the best choice for regularization, which is evident from Table~\ref{tb:reg}. In practice, these SSE probabilities, just like dropout rate, can be treated as tuning parameters and easily tuned.

\subsection{Attention Maps for Input Embeddings}
Apart from evaluating our SSE-PT against SASRec using well-defined ranking metrics on 5 datasets, we use 2 other ways to visualize the comparisons. 
The first way is to visualize the attention maps of both methods and compare them.  Note that the attention map is a lower triangular matrix as we only allow attention at present is paid to the past, but not to the future. The attention maps for the first layer in Figure~\ref{fig:attention} show that our SSE-PT paid more attention to recent items in a long sequence than SASRec. This is evident by comparing the attention intensity level of the two plots (right bottom).


As our second way to visualize the comparisons, we examine some random users' engagement histories to see the top-$K$ recommendations the two models give. In Figure~\ref{fig:example}, a random user's engagement history in Movielens1M dataset is given in temporal order (column-wise). We hide the last item whose index is 26 in test set and hope that a temporal collaborative ranking model can figure out item-26 is the one this user will watch next using only previous engagement history. One can see for a typical user, they tend to look at different style of movies at different times. Earlier on, they watched a variety of movies, including Sci-Fi, animation, thriller, romance, horror, action, comedy and adventure. But later on, in the last two columns of Figure~\ref{fig:example}, drama and thriller are the two types they like to watch most, especially the drama type. In fact, they watched 9 drama movies out of recent 10 movies. It is not surprising to see the one we hide from the models is also drama type. In the top-5 recommendations given by our SSE-PT, the hidden item-26 is put in the first place. Intelligently, the SSE-PT recommends 3 drama movies, 2 thriller movies and mixing them up in positions.  Interestingly, the top recommendation is `Othello', which like the recently watched `Richard III', is an adaptation of a Shakespeare play, and this dependence is reflected in the attention weight.  In contrast, SASRec cannot provide top-5 recommendations that are personalized enough. It recommends a variety of action, Sci-Fi, comedy, horror, and drama movies but none of them match item-26. Although this user has watched all these types of movies in the past, they do not watch these anymore as one can easily tell from his recent history. Unfortunately, SASRec cannot capture this and does not provide personalized recommendations for this user by focusing more on drama and thriller movies. What we see from this particular example is consistent with the previous findings from examining the attention maps. Attention heat maps for both models during inference are included in Figure~\ref{fig:example}. It is easy to see that SSE-PT model shares with human reasoning that more emphasis should be placed on recent movies. 




\begin{figure}[ht]
\vskip -0.1in
\begin{center}
\centerline{\includegraphics[width=0.6\columnwidth]{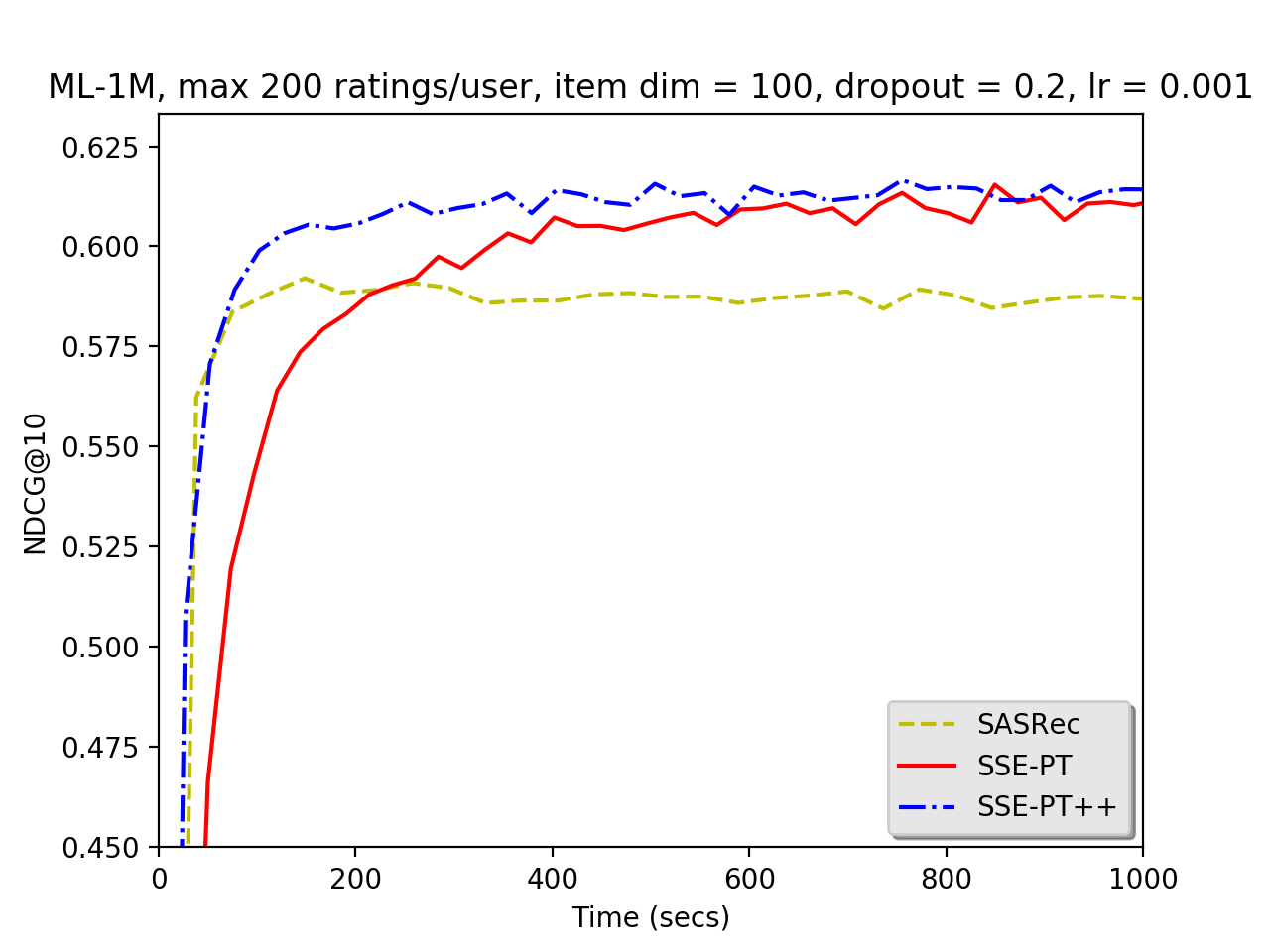}}  
\end{center}
\vskip -0.2in
\caption{Illustration of the speed of SSE-PT}
\label{fig:speed}
\end{figure}


\begin{table}[ht]
\centering
\vskip -0.1in
\caption{Comparing Different SSE probability for user embeddings for SSE-PT on Movielens1M Dataset. Embedding hidden units of 50 for users and 100 for items, attention blocks of 2, SSE probability of 0.01 for item embeddings, dropout probability of 0.2 and max length of 200 are used.}
\vskip -0.1in
\label{tb:ssep}
\begin{center}
\begin{small}
\begin{sc}
\resizebox{0.6\textwidth}{!}{
\begin{tabular}{cccc}
\toprule
 User-Side SSE-SE Probability & NDCG@$10$ & Recall@$10$ \\
\midrule
Parameter Sharing & 0.6188 & 0.8294 \\ 
\midrule
 1.0 & 0.6258  &  0.8346  \\
 0.9 & \bfseries{0.6275}  &  0.8321  \\
 0.8 & 0.6244 & 0.8359 \\
 0.6 & 0.6256 & 0.8341 \\
 0.4 & 0.6237 & \bfseries{0.8369} \\
 0.2 & 0.6163 & 0.8281 \\
 0.0 & 0.5908 & 0.8048 \\
 
\bottomrule
\end{tabular}
}
\end{sc}
\end{small}
\end{center}
\end{table}

\begin{table}[ht]
\centering
\vskip -0.1in
\caption{Comparing Different Sampling Probability, $p_s$, of SSE-PT++ on Movielens1M Dataset. Hyper-parameters the same as Table~\ref{tb:ssep}, except that the max length $T$ allowed is set 100  instead of 200 to show effects of sampling sequences.}
\vskip -0.1in
\label{tb:sampling}
\begin{center}
\begin{small}
\begin{sc}
\resizebox{0.6 \textwidth}{!}{
\begin{tabular}{cccc}
\toprule
 Sampling Probability & NDCG@$10$ & Recall@$10$ \\
\midrule
SASRec ($T=100$) & 0.5769 &  0.8045\\ 
SSE-PT ($T=100$) & 0.6142 & 0.8212  \\
\midrule
 1.0 & 0.5697   &  0.7977  \\
 0.8 &0.5735  & 0.7801  \\
 0.6 & 0.6062  & 0.8242 \\
 0.4 &  0.6113 &  0.8273 \\
 0.3 &  0.6186 & \bfseries{0.8318} \\
 0.2 & \bfseries{0.6193} & 0.8233 \\
 0.0  & 0.6142 & 0.8212  \\
 
\bottomrule
\end{tabular}
}
\end{sc}
\end{small}
\end{center}
\end{table}


\begin{table}[ht]
\centering
\vskip -0.1in
\caption{Comparing Different Number of Blocks for SSE-PT while Keeping The Rest Fixed on Movielens1M and Movielens10M Datasets.}
\vskip -0.1in
\label{tb:block}
\begin{center}
\begin{small}
\begin{sc}
\resizebox{0.6\textwidth}{!}{
\begin{tabular}{ccccc}
\toprule
Datasets & \# of blocks & NDCG@$10$ & Recall@$10$ \\
\midrule
\multirow{7}{*}{Movielens1M}
& SASREC (6 blocks) & 0.5984 & 0.8207 \\ 
\cmidrule(r){2-4} 
& 1 & 0.6162  &  0.8301  \\
& 2 & 0.6280  &  0.8365  \\
& 3 & 0.6293 & 0.8376 \\
& 4 & 0.6270 & \bfseries{0.8401} \\
& 5 & \bfseries{0.6308} & 0.8361 \\
& 6 & 0.6270 & 0.8397 \\
\midrule
\multirow{7}{*}{Movielens10M}
& SASRec (6 blocks) & 0.7531 & 0.9490 \\
\cmidrule(r){2-4} 
& 1 &  0.7454 & 0.9478    \\
& 2 &  0.7512 &  0.9522  \\
& 3 &  0.7543 & 0.9491  \\
& 4 &  0.7608 & 0.9485  \\
& 5 & 0.7619 & 0.9524  \\
& 6 & \bfseries{0.7683} & \bfseries{0.9537}  \\
\bottomrule
\end{tabular}
}
\end{sc}
\end{small}
\end{center}
\end{table}

\begin{table}[t]
\centering
\vskip -0.1in
\caption{Varying number of negatives $C$ in evaluation on Movielens1M dataset. 
Other hyper-parameters are fixed for a fair comparison. }
\vskip -0.1in
\label{tb:negative}
\begin{center}
\begin{small}
\begin{sc}
\resizebox{0.6\textwidth}{!}{
\begin{tabular}{lcccc}
\toprule
 METRIC & NDCG@$10$ & Recall@$10$ & $C$ \\
\midrule
Un-Personalized   &  0.3787 &  0.6119   & 500 \\
 Personalized  & \bfseries{0.3846} & \bfseries{0.6171}  & 500 \\
 \midrule
 Un-Personalized   &  0.2791 &  0.4781    & 1000 \\
 Personalized  & \bfseries{0.2860} & \bfseries{0.4929} & 1000 \\
 \midrule
 Un-Personalized   &  0.1939 &  0.3515    & 2000 \\
 Personalized  & \bfseries{0.1993} & \bfseries{0.3667} & 2000 \\
\bottomrule
\end{tabular}
}
\end{sc}
\end{small}
\end{center}
\end{table}


 

\subsection{Training Speeds}
In \cite{kang2018self}, it has been shown that SASRec is about 11 times faster than Caser and 17 times faster than GRU4Rec$^+$ and achieves much better NDCG@10 results so we did not include Caser and GRU4Rec$^+$ in our comparisons. Therefore, we only compare the training speeds and ranking performances among SASRec, SSE-PT and SSE-PT++.
Given that we added additional user embeddings into our SSE-PT model, it is expected that it will take slightly longer to train our model than un-personalized SASRec.
In Figure~\ref{fig:speed}, max length $T=100$ is used for SSE-PT++, and $T=200$ is used for SSE-PT and SASRec.
We find empirically that training speed of the SSE-PT and SSE-PT++ model are comparable to that of SASRec, with SSE-PT++ being the fastest and the best performing model. 
It is clear that our SSE-PT and SSE-PT++ achieve much better ranking performances than our baseline SASRec using the same training time in Figure~\ref{fig:speed}.

\subsection{Ablation Study}

\subsubsection{SSE probability} Given the importance of SSE regularization for our SSE-PT model, we carefully examined the SSE probability for input user embedding in Table~\ref{tb:ssep}. We find that the hyper-parameter SSE probability is not too sensitive: anywhere between 0.4 and 1.0 gives good results, better than parameter sharing and not using SSE-SE. This is also evident based on comparison results in Table~\ref{tb:reg}.

\subsubsection{Sampling Probability} Recall that the sampling probability is unique to our SSE-PT++ model. We show in Table~\ref{tb:sampling} using an appropriate sampling probability like 0.2-0.3 would allow it to outperform SSE-PT when the same maximum length is used.

\subsubsection{Number of Attention Blocks} We find for our SSE-PT model, a larger number of attention blocks is preferred. One can easily see in Table~\ref{tb:block}, the optimal ranking performances are achieved at $B = 4, 5$ for Movielens1M dataset and at $B = 6$ for Movielens10M dataset. 

\subsubsection{Number of Negatives Sampled} We want to make sure the number of negatives sampled during evaluation or difference in the usage of regularization techniques does not affect our final conclusion. So we add another set of experiments to remove personalization for our SSE-PT model while keeping all the regularization techniques we used. Based on the results in Table~\ref{tb:negative}, we are positive that the personalized model always outperforms the un-personalized one even if we use the same regularization techniques. This holds true regardless of how many negatives sampled during evaluations.

\section{Conclusion}
In this paper, we propose a novel neural network architecture called Personalized Transformer for the temporal collaborative ranking problem. It enjoys the benefits of being a personalized model, therefore achieving better ranking results for individual users than the current state-of-the-art. By examining the attention mechanisms during inference, the model is also more interpretable and tends to pay more attention to recent items in long sequences than un-personalized deep learning models.

\bibliography{neurips_2019}
\bibliographystyle{neurips_2019}

\end{document}